\documentclass[10pt,letterpaper]{article}

\usepackage{cogsci}

\cogscifinalcopy 
\usepackage{makecell}
\usepackage{graphicx}
\usepackage{subcaption}
\usepackage{comment}
\usepackage{todonotes}
\usepackage{booktabs}
\usepackage{adjustbox}
\usepackage{multirow}
\usepackage{array}
\usepackage{rotating}
\usepackage{enumitem}
\usepackage{siunitx}
\usepackage{microtype}

\usepackage[
  style=apa,
  natbib=true,
  annotation=false,
]{biblatex}
\usepackage{float} 

\title{Is Child-Directed Language Optimized for Word Learning? \\
A Computational Study of Verb Meaning Acquisition}

 \author{
  {\large\bfseries Francesca Padovani (f.padovani@rug.nl)}\\
  {\large\bfseries Jaap Jumelet (j.w.d.jumelet@rug.nl)}\\
  {\large\bfseries Yevgen Matusevych (yevgen.matusevych@rug.nl)}\\
  {\large\bfseries Arianna Bisazza (a.bisazza@rug.nl)}\\
   {\normalsize\normalfont
     Center for Language and Cognition (CLCG), University of Groningen \\
   }
 }

\begin{document}

\maketitle

\begin{abstract}

Is child-directed language (CDL) optimized to support language learning, and which aspects of linguistic development does it facilitate? 
We investigate this question using neural language models trained on CDL versus adult-directed language (ADL). We selectively remove syntactic or lexical co-occurrence information from the model training data, and evaluate the impact of these manipulations on verb meaning acquisition.
While disrupting syntax impairs learning across all datasets, models trained on CDL and spoken ADL show significantly higher resilience than those trained on written input. 
Tracking semantic and syntactic performance over training, we observe a “semantic-first” trajectory, with verb meanings emerging prior to robust syntactic proficiency, an asynchrony most pronounced in the spoken domain, especially CDL. These results suggest that the advantage for verb learning previously attributed to CDL may instead reflect broader properties of the spoken register, rather than a uniquely CDL-specific optimization\footnote{Repository: \url{https://github.com/fpadovani/cdl_verb_meaning}}.

\end{abstract}

\section{Introduction}

Child-directed language (CDL)\footnote{Throughout this paper, we use the term CDL specifically to refer to transcripts of child-directed speech.} is thought to provide input that is particularly conducive to learning, potentially offering advantages over adult-directed language (ADL) \citep{Ferguson1964BabyTI}. This perspective has influenced both theoretical and empirical studies on early language acquisition \citep{golinkoff,schick_function}. Recently, however, this view has been critically reassessed by \citet{kempe2024does}, who suggest that evidence for facilitative effects of CDL is largely restricted to certain linguistic features, such as prosody and register discrimination, and may not generalize to other aspects of language (e.g., morphosyntax and semantic learning). 
These observations indicate that, although some facilitative effects of CDL have been demonstrated, systematic comparisons with ADL across broader linguistic features remain limited.

Since children’s natural linguistic environments cannot be experimentally manipulated without ethical concerns, it remains difficult to directly assess how different characteristics of the input shape learning trajectories. Computational modeling has therefore emerged as a valuable tool for studying language development, providing a controlled setting in which theoretical hypotheses can be tested. Recent advances in Natural Language Processing make it possible to train large-scale language models (LMs) under different input conditions, allowing researchers to examine how specific properties of the input influence emerging linguistic patterns \citep{warstadt2022artificial,portelance1}. Reflecting the growing interest in developmentally plausible corpora, community initiatives such as the BabyLM Challenge \citep{babylm-2025-main, WILCOX2025104650} have promoted systematic investigations of how training data properties influence model behavior, with much of the work to date focusing on syntax and comparing models trained on CDL to those trained on ADL  \citep{huebner-etal-2021-babyberta,feng-etal-2024-child}. Evidence regarding the advantages of CDL for syntactic learning is inconclusive, and recent evaluations indicate that its benefits disappear when frequency effects are controlled for \citep{padovani-etal-2025-child}. 

Beyond syntax, semantic development constitutes another central component of early language acquisition, and verb learning in particular poses a well-known challenge due to the abstract and relational nature of verb meaning \citep{gentner1978relational}. This study therefore investigates whether CDL is structured to facilitate the acquisition of verb meaning.
This issue has been previously examined by \citet{you2021child}, who argue that the statistical properties of English CDL, relative to both spoken and written ADL, support the acquisition of verb categories without reliance on high-level syntactic knowledge. On this basis, they propose that CDL may enable syntax-independent semantic inference, challenging the \textit{syntactic bootstrapping} hypothesis \citep{gleitman1990structural}.
However, their analysis is limited both in scope, focusing only on the distinction between causative and non\-causative verbs, and in coverage, as it includes only a small set of verbs for each category. Combined with the use of a non-contextual distributional model, these constraints raise questions about the generalizability of their conclusions to other verbs and to other types of statistical learners. More recently, \citet{zhu2025structural}, also working on English, adopt a complementary perspective by assuming a central role for syntactic bootstrapping in children, and testing whether contextual Transformer-based LMs trained on CDL exhibit similar learning behavior to humans. By manipulating the input data prior to training, they show that altering syntactic structure impairs verb meaning acquisition more than does perturbing lexical co-occurrences. These results suggest that LMs, like human learners, rely on syntactic cues to infer verb meaning. However, \citeauthor{zhu2025structural}'s analysis is limited to CDL, leaving open the question of whether this type of language is more robust to syntax degradation when acquiring verb meaning, compared to other input domains.

In the present study, we address this gap by explicitly comparing CDL to both spoken and written ADL, being guided by three main research questions:

\begin{itemize}
    \item \textbf{RQ1:} Does perturbing syntactic structure affect verb meaning acquisition more than perturbing lexical co-occurrence only in CDL or also in ADL?
    \item \textbf{RQ2:} Is verb meaning acquisition in CDL, compared to ADL, more robust to the ablation of syntax?
    \item \textbf{RQ3:} How do the developmental trajectories of verb meaning acquisition and syntactic performance relate over time? How do these trajectories differ between CDL and ADL?
\end{itemize}

\noindent
To address these questions, we train autoregressive Transformer LMs and use the data manipulation framework of \citet{zhu2025structural}, applying controlled perturbations to the training corpora (\textbf{RQ1, RQ2}). Verb meaning acquisition is evaluated with in-domain semantic minimal pairs, and its relation to syntactic development is assessed using established syntactic minimal pair benchmarks (\textbf{RQ3}).

\section{Related Work}
\subsubsection{The Verb Learning Problem}
Understanding how children acquire verb meaning has long been a central problem in language acquisition: unlike nouns, verbs encode relational meaning and require learners to infer abstract event structures from sparse and noisy linguistic input \citep{gentner1978relational}. Much work emphasizes the role of syntax in constraining verb interpretation, an idea formalized in the syntactic bootstrapping hypothesis \citep{gleitman1990structural, babineau2024syntactic}. Support for this view comes, among others, from \citet{GILLETTE1999135}, who show that verbs are more readily identified when learners have access to syntactic frames than to cross-situational cues alone. Other studies demonstrate that learners can extract aspects of verb meaning from contextual regularities in the input, even in the absence of fully developed syntactic representations \citep{newport2004learning,smith2008infants}. Empirical evidence suggests that children flexibly integrate multiple sources of information during verb learning, including word co-occurrence statistics and syntactic frames \citep{naigles1996use, lidz2003infants}. 

\subsubsection{Assessing Linguistic Performance in LMs}
Inspired by language acquisition research, recent work in Computational Linguistics has examined how lexical meaning emerges in LMs trained on developmentally plausible input, such as CHILDES transcripts \citep{macwhinney2000childes}. A common category of evaluation, surprisal-based approaches, assesses word learning by evaluating a model's ability to anticipate words in context: words that are highly predictable in a given context elicit lower surprisal, reflecting the model’s confidence in its predictions
\citep{chang-bergen-2022-word, portelance2023predicting, shafiabadi-wisniewski-2025-beyond}. However, such measures have limitations: they show when words are predicted reliably but not which cues the model uses to learn them. Moreover, even frequent words can appear surprising in unusual contexts, since surprisal mainly reflects the next-word prediction objective rather than abstract word knowledge. Complementary to surprisal-based approaches, lexical decision paradigms probe whether models distinguish real words from plausible non-words as a proxy for semantic performance \citep{goriely-etal-2024-babble, bunzeck-zarriess-2025-subword}. Minimal pairs offer another well-established paradigm for assessing linguistic knowledge in LMs. Based on the observation that small changes can render a sentence acceptable or unacceptable to native speakers \citep{chomsky2014aspects}, minimal pairs test whether a model assigns higher probability to the grammatical or contextually appropriate sentence when only a single element differs \citep{linzen-etal-2016-assessing, marvin-linzen-2018-targeted}. Benchmarks like BLiMP \citep{warstadt-etal-2020-blimp-benchmark}, Zorro \citep{huebner-etal-2021-babyberta} and CLAMS \citep{mueller-etal-2020-cross} primarily evaluate syntax, whereas semantic minimal pairs assess whether models prefer the contextually or semantically correct verb over an alternative \citep{zhu2025structural}.

\subsubsection{Computational Studies of Verb Learning Mechanisms}
Two recent studies are particularly inspirational for this work. 
\begin{itemize}[leftmargin=0pt, labelwidth=*, align=left]
\item 
\citet{you2021child} (henceforth \textbf{Y21}) test whether distributional statistics in child-directed language suffice for verb meaning acquisition. They train Word2Vec models \citep{mikolov}, non-contextual distributional models, on CDL and on spoken and written ADL. Focusing on causative versus non-causative verbs, they find that models trained on CDL distinguish these verb classes more accurately than models trained on ADL, despite not having access to explicit syntactic structure. 
This discriminative power is strongest in smaller context windows, suggesting that CDL’s repetitive and predictable patterns provide dense, redundant cues helping learners infer semantic regularities from the input. Although limited to a small set of verbs (23 causative + 9 non-causatives), their conclusion aligns with \citet{futrell-2025-language}, who emphasizes the critical role of local redundancy in language learning and highlights how distributional statistics can support the learning of form–meaning mappings, even when learners only have access to surface forms. 
Relatedly, \citet{lee2025readability} also show that model learnability is predicted by statistical simplicity, operationalized as reduced \textit{n}-gram complexity, highlighting the importance of regular and redundant input patterns for language modeling. We adopt from Y21 the comparative setup that contrasts CDL with spoken and written ADL. Additionally, their findings motivate our hypothesis that CDL may be especially well suited to support verb learning even when syntactic cues are limited.

\item 
\citet{zhu2025structural} (henceforth \textbf{Z25}) examine whether LMs exhibit syntactic bootstrapping by evaluating verb meaning acquisition across all verb forms in the training data, using both RoBERTa \citep{liu2019roberta}, a masked LM, and GPT-2 \citep{radford2019language}, a causal LM. Through targeted manipulations of word order and lexical co-occurrence in CDL, Z25 shows that disrupting syntactic cues leads to larger impairments in verb learning than does perturbing distributional statistics, providing evidence for syntactic bootstrapping in LMs trained on this domain. Because their analysis is restricted to CDL, it remains unclear how the relative importance of syntactic cues varies across CDL and spoken and written ADL. We adopt from Z25 the methodological framework 
involving the selective ablation of syntactic and co-occurrence information, to systematically probe verb meaning acquisition across different input domains.
\end{itemize}

\section{Experimental Setup}
We assess verb meaning acquisition in English by comparing Tranformer-based LMs trained on CDL to those trained on three additional corpora. Following Z25, we train models on both original and manipulated versions of each corpus. Verb meaning acquisition is evaluated via in-domain semantic minimal pairs for each dataset. Finally, we examine how verb learning relates to the emergence of syntactic knowledge by testing models on standard syntactic minimal pair benchmarks \citep{warstadt-etal-2020-blimp-benchmark,huebner-etal-2021-babyberta, padovani-etal-2025-child}.

\subsection{Training Data}
We follow Y21 in using CDL and the spoken and written portions of BNC \citep{davies2004byu}, while also including Wikipedia to provide a complementary written domain with encyclopedic prose, extending the analysis beyond the mostly narrative content of the BNC. We base the size of all datasets on the amount of data available in the CDL corpus, which, after preprocessing, comprises a total of 15M tokens (10/2.5/2.5M for training/dev/test set, respectively). All other datasets are split according to the same train/dev/test proportions.

\subsubsection{Child-Directed Language}
Following Z25, we use developmentally plausible CDL data from \textit{CHILDES} \citep{macwhinney2000childes}, specifically the CHILDES-only portion of the BabyLM Challenge corpus\footnote{\url{https://babylm.github.io/}}
\citep{babylm-2025-main}. The raw corpus contains roughly 29M tokens; after removing speaker tags and paralinguistic material, 15M tokens remain.

\subsubsection{ADL Written -- British National Corpus}
We extract a portion of the written part of the \textit{British National Corpus} (BNC) \citep{davies2004byu}. The BNC comprises 100M words spanning a wide range of genres, including spoken language, fiction, magazines, newspapers, and academic texts. We analyze the distribution of documents in the written portion, observing a predominance of books, periodicals, and TV news. 

\subsubsection{ADL Written -- Wikipedia}  
\textit{Wikipedia} is a collection of written articles that have an encyclopedic format, and it is commonly used as a standard resource for training LMs. We use the existing Wikipedia splits from \citet{huebner-etal-2021-babyberta}.\footnote{\url{https://github.com/phueb/BabyBERTa}} 

\subsubsection{ADL Spoken -- CANDOR + Switchboard + BNC}
To represent adult conversational language, we construct a spoken-domain dataset using \textit{CANDOR}\footnote{\url{https://guscooney.com/candor-dataset/}} \citep{reece2023candor}, collected from video-call recordings between strangers during the 2020 pandemic. As CANDOR contains fewer than 15M tokens, we supplement it with the \textit{Switchboard Dialog Act Corpus} \citep{stolcke2000dialogue} and a sample from the spoken portion of the \textit{BNC} \citep{davies2004byu}, achieving the target dataset size.

\vspace{0.5mm}
Table~\ref{tab:dataset-stats} summarizes key statistics of the training datasets. As expected, the written corpora exhibit higher type-token ratios and longer average sentence lengths than the spoken datasets, reflecting greater lexical diversity and syntactic complexity.

\begin{table}[t]
\footnotesize
\begin{center} 
\setlength{\abovecaptionskip}{4pt}  
\setlength{\belowcaptionskip}{2pt}  
\caption{Descriptive statistics of the training datasets. TTR refers to type-token ratio.}
\label{tab:dataset-stats}
\begin{tabular}{l S S S S} 
\hline
\hline
\noalign{\vskip 2mm}
Statistic & {CDL} & {CANDOR} & {BNC} & {Wikipedia} \\ [2mm]
\hline
\hline
\noalign{\vskip 2mm}
1-gram TTR       & 0.004 & 0.005 & 0.013 & 0.021\\
2-gram TTR      & 0.109 & 0.124 &  0.269 & 0.309\\
3-gram TTR      & 0.420 & 0.442 & 0.684 & 0.715\\
Avg sent length  & 4.55 & 9.31  & 20.40 & 20.34\\ 
\hline
\end{tabular} 
\end{center} 
\end{table}

\subsection{Training Data Manipulations}

To investigate how syntactic and lexical co-occurrence information contribute to verb meaning acquisition, we apply targeted perturbations to the training data following the framework of Z25.\footnote{The original Z25 code was not publicly available at the time of this work, so we independently implemented the perturbation pipeline.} We then compare the effects of these manipulations across CDL and ADL to assess whether the patterns observed in CDL generalize or diverge in other domains. Below, we describe both conditions in detail:

\subsubsection{REPLACE.WORD – co-occurrence disruption}
 Under this manipulation, lexical items that co-occur with verbs, specifically nouns, adjectives, and adverbs, are replaced within each sentence by other words with the same part-of-speech and fine-grained XPOS tag (i.e., language-specific morphological tags encoding features such as number, tense, or degree). Replacement words are sampled according to frequency bins for the corresponding POS and XPOS tags, thus preserving the overall token frequency distribution. In sentences containing multiple verbs, the root verb is left unchanged, while dependent verbs are replaced following the same procedure as for other parts of speech. This manipulation preserves surface word order and syntactic structure while selectively removing verb-specific distributional cues. The proportion of replaced words is 10.3\% for CDL, 11.3\% for CANDOR, 26.3\% for Wikipedia, and 24.2\% for BNC, reflecting differences in average sentence length across domains (Table~\ref{tab:dataset-stats}). This manipulation contributes to \textbf{RQ1} and is used in combination with SHUFFLE.ORDER to test whether the relative impact of co-occurrence versus syntactic disruptions observed in CDL generalizes to ADL.

\subsubsection{SHUFFLE.ORDER -- word order disruption}
This simple manipulation exploits the fact that English encodes syntactic structure primarily through word order. Words are randomly shuffled within each sentence, eliminating linear order dependencies and degrading syntactic information while retaining sentence-level co-occurrence statistics. This manipulation directly operationalizes our \textbf{RQ2}: by disrupting word order within sentences in CDL vs.\ ADL we can assess whether verb meaning acquisition is more resilient in CDL.

\subsection{Models and Training} We focus on autoregressive LMs, specifically GPT-2 \citep{radford2019language}, due to the higher cognitive plausibility of their training objective compared to bidirectional models. We follow the setup of Z25, using a model architecture consisting of 12 layers, 12 attention heads per layer, and 768 hidden units. Training is conducted with a linear learning rate scheduler, a learning rate of $10^{-4}$, and a batch size of 256. To examine the joint development of semantic and syntactic performance over an extended learning period (\textbf{RQ3}), models are trained for 20 epochs. For each dataset and experimental condition, results are averaged across five random initializations (seeds). Results reported for \textbf{RQ1} and \textbf{RQ2} are based on the checkpoint with the lowest validation loss on the development set for each corpus and experimental condition, as mild overfitting was typically observed after several training epochs.

\subsection{Evaluation}
We adopt a minimal pair paradigm commonly used to probe linguistic knowledge in LMs. Each test item consists of a pair of sentences that differ only in one targeted property, while all other aspects are held constant. A model is considered to make a correct prediction if it assigns a higher probability to the grammatical or contextually appropriate sentence than to its minimally altered counterpart. Sentence probabilities are computed using the \texttt{minicons} library \citep{misra2022minicons}.

\subsubsection{Semantic Minimal Pairs}
Following Z25, we generate in-domain semantic minimal pairs from the test portion of each dataset. Since we focus on verb meaning acquisition, we identify the target verb (i.e., the root verb) in each sentence and create up to five alternative sentences by substituting verbs from the same fine-grained frequency bin, defined over POS and XPOS categories from the training data. To ensure comparability across domains, we limit sentence length to 10–30 tokens. This results in minimal pairs that differ only in the target verb, isolating the contribution of lexical–semantic appropriateness.\footnote{Two methodological considerations apply: first, some frequency bins are sparse and do not always contain five suitable replacement verbs, leading to fewer than five generated minimal pairs. Second, this procedure does not guarantee that every alternative is strictly unacceptable; nevertheless, substituting the target verb with a contextually mismatched alternative should produce less plausible sentences on average, even if not at the individual level.}
In total, we obtain four distinct sets, one per training corpus, each containing 140,000 minimal pairs and spanning a broad range of verb types (between  801 and 2,765 unique verb lemmas across datasets).
 
An example is provided in \ref{ex:1}.

\begin{enumerate}[label={(\arabic*)}, series=cogsci-ex, leftmargin=2.5em]
    \footnotesize
    \item \label{ex:1}
    \begin{enumerate}[label=\alph*., leftmargin=1.5em, nosep]
        \item You can sit out here by me on the other stool.
        \item * You can \textit{try} out here by me on the other stool.
    \end{enumerate}
\end{enumerate}

\noindent
As a sanity check, we verify that our semantic minimal pairs capture genuine verb meaning knowledge rather than dataset-specific artifacts. We use minimal pairs generated from one dataset (e.g., CDL) to test models trained on another (e.g., BNC). Models achieve high accuracy only when training and test domains match, showing that strong in-domain performance reflects true acquisition of domain-specific verb patterns rather than quirks of the evaluation procedure.
All results reported in the Results section refer to in-domain train/test setup.

\subsubsection{Syntactic Minimal Pairs}
To test models across a range of syntactic phenomena, we first consider two benchmarks commonly used for English LMs: BLiMP \citep{warstadt-etal-2020-blimp-benchmark} and Zorro \citep{huebner-etal-2021-babyberta}. However, \citet{padovani-etal-2025-child} recently highlighted the limitations of these benchmarks when applied to language with distinct lexical distributions or domain-specific vocabularies, such as CDL. As a solution, they proposed to generate corpus-specific test sets, ensuring the minimal pairs contain words drawn from comparable frequency ranges in each corpus. 
They applied this method to CLAMS \citep{mueller-etal-2020-cross}, resulting in the FIT-CLAMS benchmark, which we adopt for our analysis. CLAMS comprises only the syntactic paradigm of subject–verb number agreement, tested across different clause types (e.g., simple clauses, prepositional phrases, coordinates, and relative clauses).
As the original FIT-CLAMS versions were developed for CDL and Wikipedia,
here we also apply the same procedure to generate two new versions, for BNC and CANDOR. 
This ensures that syntactic performance can be probed under comparable conditions across all four datasets. Example~\ref{ex:2} illustrates a syntactic minimal pair testing \textit{subject--verb number agreement}:

\begin{enumerate}[label={(\arabic*)}, resume=cogsci-ex, leftmargin=2.5em]
    \footnotesize
    \item \label{ex:2}
    \begin{enumerate}[label=\alph*., leftmargin=1.5em, nosep]
        \item The painter in front of the waiter enjoys.
        \item * The \textit{painters} in front of the waiter enjoys.
    \end{enumerate}
\end{enumerate}

In the next section, we report results for \textbf{RQ1} and \textbf{RQ2}, evaluating models trained under different data manipulations on semantic minimal pairs, and for \textbf{RQ3}, tracking accuracy on semantic and syntactic minimal pairs over time for models trained on the original datasets.
\section{Results}
\subsubsection{RQ1: Effect of Co-Occurrence Disruptions on Verb Learning}
\begin{figure}[t]
\includegraphics[width=0.93\columnwidth]{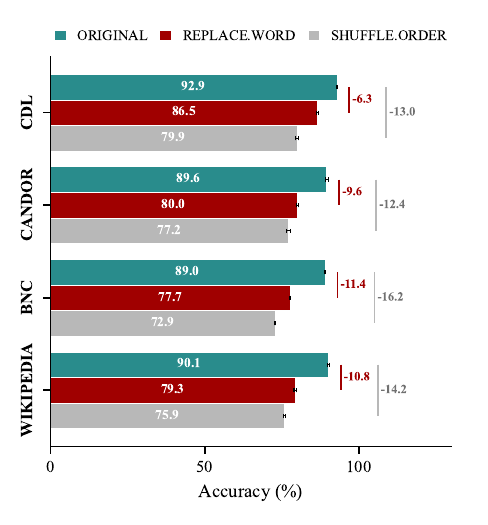}
\captionsetup{skip=0pt} 
\caption{Accuracy scores on semantic minimal pairs in the three different experimental conditions for each data domain.} 
\label{fig:small-barplot}
\end{figure}

We first replicate Z25's finding that models trained on CDL are more sensitive to syntactic disruptions than to changes in lexical co-occurrence when evaluated on semantic minimal pairs targeting verb meaning. Perturbing word order produces a larger drop in accuracy than does altering verb co-occurrences, suggesting that LMs, like human learners, rely on syntactic scaffolding to infer verb meaning. We then extend this analysis to the other three corpora to examine whether the relative impact of syntactic versus lexical disruptions observed in CDL generalizes to adult-directed input. As shown in Figure~\ref{fig:small-barplot}, the accuracy drop from ORIGINAL to REPLACE.WORD (red) is consistently smaller than the drop from ORIGINAL to SHUFFLE.ORDER (gray), not only in CDL but also in spoken (CANDOR) and written ADL (BNC, Wikipedia). While this differential impact is shared across domains, the magnitude of the drop caused by lexical perturbations varies with corpus properties. In particular, the accuracy drop under this manipulation scales with average sentence length (reported in Table~\ref{tab:dataset-stats} above) and with the lexical replacement rates as reported in the REPLACE.WORD manipulation section. Intuitively, longer sentences with more words co-occurring with the main verb lead to larger drops under lexical perturbations.
Overall, these results answer \textbf{RQ1} by showing that syntactic cues provide critical scaffolding for verb learning in both CDL and ADL, while lexical co-occurrences exert a more limited influence. 

\subsubsection{RQ2: Effect of Syntax Perturbation on Verb Learning}
\textbf{RQ2} directly builds on the results of \textbf{RQ1} and on the observation that syntax plays a key role in verb meaning acquisition across all data domains. Drawing on Y21's claim that CDL may be optimized for syntax-free semantic inference, we ask whether the impact of syntactic perturbations differs across domains and whether CDL shows greater resilience to word order disruptions than written or spoken ADL. While we expect CDL to maintain relatively high accuracy due to its repetitive and predictable input, we do not make strong predictions for adult-directed spoken (CANDOR) or written (Wikipedia, BNC) corpora, although the greater syntactic complexity of written text may make it more sensitive to word order manipulations. To formally evaluate the robustness of verb learning when syntax is ablated, we employed an Ordinary Least Squares (OLS) regression model\footnote{We initially tested a mixed-effects model with seed as a random effect, but the random effect variance was effectively zero, so we used a standard linear regression instead.} to predict accuracy as a function of Dataset (CDL, CANDOR, BNC, Wikipedia), Condition (ORIGINAL vs. SHUFFLE.ORDER), and their interaction. By setting the CDL baseline in the ORIGINAL condition as our reference point (the intercept), we could precisely quantify the performance decrement associated with syntactic disruption and determine whether the drop in accuracy varied significantly by corpus. As summarized in Figure~\ref{fig:small-barplot}, models trained on CDL achieve high baseline performance (92.9\%), which decreases by 13.0\% when word order is corrupted, a significant main effect of Condition ($\beta = -0.13$, $p < .001$). Crucially, the interaction terms reveal that the impact of syntactic ablation is not uniform across data domains. The accuracy drop is significantly more pronounced in the written corpora than in CDL, with BNC and Wikipedia showing additional decreases of 3.2\% ($\beta = -0.032$, $p < .001$) and 1.2\% ($\beta = -0.012$, $p < .001$), respectively. By contrast, the interaction term for the adult-directed spoken corpus (CANDOR) is not significant ($\beta = 0.005$, $p = .088$). These results suggest that the robustness to syntactic perturbation is not exclusive to CDL. Rather, the comparable performance of CANDOR indicates that a reduced reliance on word order may be a general characteristic of learning from conversational speech, contrasting with the higher sensitivity to structural changes in representations derived from written text.

\subsubsection{RQ3: Developmental Trajectories of Semantics and Syntax}
\begin{figure}[!t]  
\centering
\includegraphics[width=0.9\columnwidth]{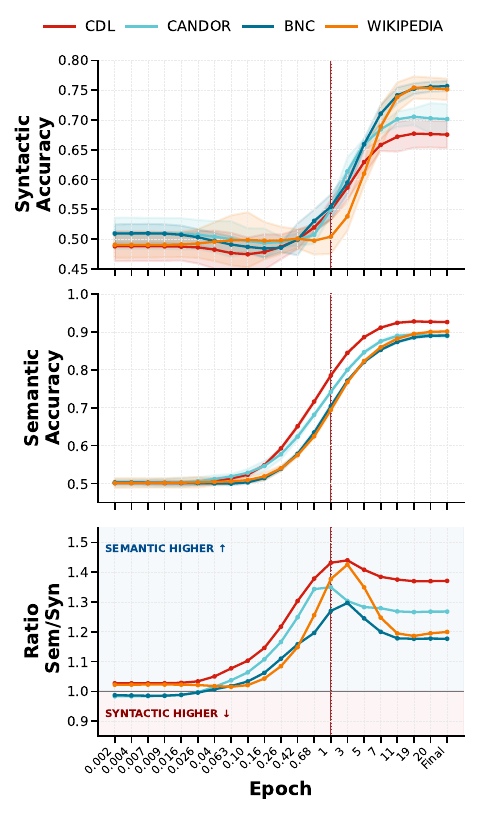}
\captionsetup{skip=0pt}
\caption{Evolution of model performance across training checkpoints for the four datasets considered in this study. Top: syntactic accuracy on FIT-CLAMS. Middle: semantic accuracy on semantic minimal pairs. Bottom: ratio of semantic to syntactic accuracy.}
\label{fig:combined-benchmarks}
\end{figure}

To address \textbf{RQ3}, we examine the evolution of models' performances over the course of training, comparing the emergence of syntactic proficiency with the acquisition of verb meaning. Our hypothesis is that the distributional properties of CDL allow models to form robust representations of verb meaning early in training, even before a sharp increase in syntactic knowledge is observed. In contrast, we expect that in more complex adult-directed domains, verb meaning development depends more strongly on—and thus progresses alongside—emerging syntactic proficiency. 
We focus our analysis on the FIT-CLAMS benchmark, which provides the most reliable cross-domain comparisons. However, we also test our models on BLiMP and Zorro, finding consistent overall patterns. The learning trajectories shown in Figure~\ref{fig:combined-benchmarks} reveal a ``semantic-first'' trend across all domains: Semantic Accuracy (middle panel) rises steadily in early epochs (0.063 to~1), while Syntactic Accuracy remains near chance. However, the datasets differ in the degree of synchrony between these two competencies. This is explicitly captured by the Ratio Sem/Syn (bottom panel), where a higher ratio indicates a greater developmental lead for semantics over syntax. CDL demonstrates the most pronounced developmental asynchrony; it maintains the highest ratio of semantic-to-syntactic knowledge from Epoch 0.04 through to the final checkpoint. By Epoch 1 (vertical line in Figure~\ref{fig:combined-benchmarks}), CDL reaches almost 80\% semantic accuracy, while syntactic performance remains under 60\%. By contrast, written ADL (Wikipedia and BNC) shows a much tighter coupling of these skills. After a brief early semantic lead, the ratio values for these two datasets drop steeply after Epoch 3 as syntactic proficiency rapidly ``catches up'' to semantic levels, resulting in a more balanced final profile. Interestingly, CANDOR occupies a middle ground, highlighting the distinction between spoken and written input. While CANDOR aligns with the CDL trajectory early on, showing a similarly high semantic lead, it eventually diverges as its syntactic acquisition accelerates, landing between the ``coupled'' written corpora and the ``decoupled'' CDL. Overall, we observe that different training domains are associated with distinct developmental trajectories. Throughout the learning process, CDL consistently maintains the highest ratio of semantic-to-syntactic knowledge. While this trend aligns with our hypothesis that CDL is optimized for verb meaning that relies less on formal syntactic proficiency, we remain cautious about drawing causal conclusions. These observations remain tied to our specific operationalization of syntactic acquisition, a limitation we discuss below.

\section{Discussion and Conclusion}
Across all domains, our results indicate that verb meaning acquisition depends more strongly on syntactic structure than on lexical co-occurrence. Importantly, the relative resilience to syntactic disruption observed for CDL does not appear to be unique to child-directed input: models trained on conversational language, including both CDL and spoken ADL (CANDOR), exhibit greater stability than those trained on written text. This pattern suggests that differences in robustness to syntactic perturbations are better explained by spoken versus written input, rather than CDL versus ADL specifically, providing a more nuanced perspective compared to \citet{you2021child}, who emphasize CDL as uniquely optimized for verb learning. Longitudinally, we observe a ``semantic-first'' trajectory where verb meanings emerge before robust syntactic proficiency across all domains, but with systematic differences in how tightly semantic and syntactic development become coupled over time. This developmental asynchrony is most pronounced within the spoken domain, particularly in CDL. Together, these findings suggest that, while syntax serves as a universal scaffold for verb learning, conversational input—whether child- or adult-directed—may support more robust semantic representations even when syntactic knowledge remains limited. 

Nevertheless, our study is limited by its coarse-grained approach to syntax, treating it primarily as word-order preservation within an English-centric context. This coarse-grained approach leaves open which specific syntactic frames or cues drive verb learning—for instance, whether transitive verbs require only correct agent--patient ordering or additional contextual signals. Extending these manipulations, e.g., by selectively scrambling different constituents or preserving local hierarchical structures, would enable a more fine-grained causal investigation of the sources of syntactic bootstrapping in language models. 

Methodologically, tracking semantic and syntactic trajectories opens avenues for analyses beyond behavioral probing. Future work could integrate interpretability methods, e.g., monitoring the emergence of Syntactic Attention Structure (SAS) \citep{chen2024sudden}, as a complementary assessment of syntactic knowledge. Utilizing such model internals could help overcome limitations of minimal pair benchmarks by providing a more direct view of how structural dependencies are encoded and emerge over time. Finally, extending this framework to typologically diverse languages is crucial to determine whether the observed patterns reflect universal properties or are specific to English.
\section{Acknowledgements}
This publication is part of the project ‘Polyglot Machines’ (VI.Vidi.221C.009) funded by the Talent Programme of the Dutch Research Council (NWO). The authors thank the members of the InCLow and GroNLP groups at the University of Groningen for their useful feedback.

\printbibliography

@article{reece2023candor,
  title={{The CANDOR corpus: Insights from a large multimodal dataset of naturalistic conversation}},
  author={Reece, Andrew and Cooney, Gus and Bull, Peter and Chung, Christine and Dawson, Bryn and Fitzpatrick, Casey and Glazer, Tamara and Knox, Dean and Liebscher, Alex and Marin, Sebastian},
  journal={Science Advances},
  volume={9},
  number={13},
  pages={eadf3197},
  year={2023},
  publisher={American Association for the Advancement of Science}
}

@article{stolcke2000dialogue,
  title={Dialogue act modeling for automatic tagging and recognition of conversational speech},
  author={Stolcke, Andreas and Ries, Klaus and Coccaro, Noah and Shriberg, Elizabeth and Bates, Rebecca and Jurafsky, Dan and Taylor, Paul and Martin, Rachel and Van Ess-Dykema, Carol and Meteer, Marie},
  journal={{Computational Linguistics}},
  volume={26},
  number={3},
  pages={339--374},
  year={2000}
}

@misc{macwhinney2000childes,
  title={{The CHILDES project: Tools for analyzing talk: Volume I: Transcription format and programs, volume II: The database}},
  author={MacWhinney, Brian},
  year={2000},
  publisher={MIT Press One Rogers Street, Cambridge, MA 02142-1209, USA journals-info~…}
}

@inproceedings{huebner-etal-2021-babyberta,
    title = "{B}aby{BERT}a: Learning More Grammar With Small-Scale Child-Directed Language",
    author = "Huebner, Philip A.  and
      Sulem, Elior  and
      Cynthia, Fisher  and
      Roth, Dan",
    editor = "Bisazza, Arianna  and
      Abend, Omri",
    booktitle = {{Proceedings of the 25th Conference on Computational Natural Language Learning}},
    year = "2021",
    address = "Online",
    publisher = "Association for Computational Linguistics",
    url = "https://aclanthology.org/2021.conll-1.49/",
    doi = "10.18653/v1/2021.conll-1.49",
    pages = "624--646"
}

@article{radford2019language,
  title={Language models are unsupervised multitask learners},
  author={Radford, Alec and Wu, Jeffrey and Child, Rewon and Luan, David and Amodei, Dario and Sutskever, Ilya},
  journal={OpenAI blog},
  volume={1},
  number={8},
  pages={9},
  year={2019}
}

@article{misra2022minicons,
  title={minicons: Enabling flexible behavioral and representational analyses of transformer language models},
  author={Misra, Kanishka},
  journal={arXiv preprint arXiv:2203.13112},
  year={2022}
}

@book{chomsky2014aspects,
  title={Aspects of the Theory of Syntax},
  author={Chomsky, Noam},
  number={11},
  year={2014},
  publisher={MIT press}
}

@article{linzen-etal-2016-assessing,
    title = "Assessing the Ability of {LSTM}s to Learn Syntax-Sensitive Dependencies",
    author = "Linzen, Tal  and
      Dupoux, Emmanuel  and
      Goldberg, Yoav",
    journal = "Transactions of the Association for Computational Linguistics",
    volume = "4",
    year = "2016",
    address = "Cambridge, MA",
    publisher = "MIT Press",
    url = "https://aclanthology.org/Q16-1037/",
    doi = "10.1162/tacl_a_00115",
    pages = "521--535"
}

@inproceedings{marvin-linzen-2018-targeted,
    title = "Targeted Syntactic Evaluation of Language Models",
    author = "Marvin, Rebecca  and
      Linzen, Tal",
    editor = "Riloff, Ellen  and
      Chiang, David  and
      Hockenmaier, Julia  and
      Tsujii, Jun{'}ichi",
    booktitle = {{Proceedings of the 2018 Conference on Empirical Methods in Natural Language Processing}},
    year = "2018",
    address = "Brussels, Belgium",
    publisher = "Association for Computational Linguistics",
    url = "https://aclanthology.org/D18-1151/",
    doi = "10.18653/v1/D18-1151",
    pages = "1192--1202"
}

@inproceedings{futrell-2025-language,
    title = {{Language Learning as Codebreaking: The Key Roles of Redundancy and Locality}},
    author = "Futrell, Richard",
    editor = "Anderson, Carolyn Jane  and
      Mailhot, Fr{\'e}d{\'e}ric  and
      Prasad, Grusha",
    booktitle = {{Proceedings of the Society for Computation in Linguistics 2025}},
    year = "2025",
    address = "Eugene, Oregon",
    publisher = "Association for Computational Linguistics",
    url = "https://aclanthology.org/2025.scil-1.5/",
    pages = "54--63"
}

@article{warstadt-etal-2020-blimp-benchmark,
    title = "{BL}i{MP}: The Benchmark of Linguistic Minimal Pairs for {E}nglish",
    author = "Warstadt, Alex  and
      Parrish, Alicia  and
      Liu, Haokun  and
      Mohananey, Anhad  and
      Peng, Wei  and
      Wang, Sheng-Fu  and
      Bowman, Samuel R.",
    journal = "Transactions of the Association for Computational Linguistics",
    volume = "8",
    year = "2020",
    address = "Cambridge, MA",
    publisher = "MIT Press",
    url = "https://aclanthology.org/2020.tacl-1.25/",
    pages = "377--392"
}

@inproceedings{mueller-etal-2020-cross,
    title = "Cross-Linguistic Syntactic Evaluation of Word Prediction Models",
    author = "Mueller, Aaron  and
      Nicolai, Garrett  and
      Petrou-Zeniou, Panayiota  and
      Talmina, Natalia  and
      Linzen, Tal",
    editor = "Jurafsky, Dan  and
      Chai, Joyce  and
      Schluter, Natalie  and
      Tetreault, Joel",
    booktitle = {{Proceedings of the 58th Annual Meeting of the Association for Computational Linguistics}},
    year = "2020",
    address = "Online",
    publisher = "Association for Computational Linguistics",
    url = "https://aclanthology.org/2020.acl-main.490/",
    doi = "10.18653/v1/2020.acl-main.490",
    pages = "5523--5539"
}

@inproceedings{padovani-etal-2025-child,
    title = "Child-Directed Language Does Not Consistently Boost Syntax Learning in Language Models",
    author = "Padovani, Francesca  and
      Jumelet, Jaap  and
      Matusevych, Yevgen  and
      Bisazza, Arianna",
    editor = "Christodoulopoulos, Christos  and
      Chakraborty, Tanmoy  and
      Rose, Carolyn  and
      Peng, Violet",
    booktitle = {{Proceedings of the 2025 Conference on Empirical Methods in Natural Language Processing}},
    year = "2025",
    address = "Suzhou, China",
    publisher = "Association for Computational Linguistics",
    url = "https://aclanthology.org/2025.emnlp-main.999/",
    doi = "10.18653/v1/2025.emnlp-main.999",
    pages = "19746--19767",
    ISBN = "979-8-89176-332-6"
}

@article{portelance1,
author = {Portelance, Eva and Jasbi, Masoud},
title = {The Roles of Neural Networks in Language Acquisition},
journal = {Language and Linguistics Compass},
volume = {18},
number = {6},
pages = {e70001},
doi = {https://doi.org/10.1111/lnc3.70001},
eprint = {https://compass.onlinelibrary.wiley.com/doi/pdf/10.1111/lnc3.70001},
year = {2024}
}

@article{babineau2024syntactic,
  title={Syntactic bootstrapping as a mechanism for language learning},
  author={Babineau, Mireille and Barbir, Monica and de Carvalho, Alex and Havron, Naomi and Dautriche, Isabelle and Christophe, Anne},
  journal={Nature Reviews Psychology},
  volume={3},
  number={7},
  pages={463--474},
  year={2024},
  publisher={Nature Publishing Group US New York}
}

@article{smith2008infants,
  title={Infants rapidly learn word-referent mappings via cross-situational statistics},
  author={Smith, Linda and Yu, Chen},
  journal={Cognition},
  volume={106},
  number={3},
  pages={1558--1568},
  year={2008},
  publisher={Elsevier}
}

@article{zhu2025structural,
  title={The Structural Sources of Verb Meaning Revisited: Large Language Models Display Syntactic Bootstrapping},
  author={Zhu, Xiaomeng and McCoy, R Thomas and Frank, Robert},
  journal={arXiv preprint arXiv:2508.12482},
  year={2025}
}

@inproceedings{lee2025readability,
  title = {Readability $\neq$ Learnability: Rethinking the Role of Simplicity in Training Small Language Models},
  author = {Lee, Ivan and Berg-Kirkpatrick, Taylor},
  booktitle = {Conference on Language Modeling (COLM)},
  year = {2025},
  publisher = {OpenReview}
}

@inproceedings{mikolov,
  author       = {Tom{\'{a}}s Mikolov and
                  Kai Chen and
                  Greg Corrado and
                  Jeffrey Dean},
  title        = {{Efficient Estimation of Word Representations in Vector Space}},
  booktitle    = {{1st International Conference on Learning Representations, {ICLR} 2013, Workshop Track Proceedings}},
  year         = {2013},
  crossref     = {DBLP:conf/iclr/2013w},
  url          = {http://arxiv.org/abs/1301.3781},
  bibsource    = {dblp computer science bibliography, https://dblp.org}
}

@proceedings{babylm-2025-main,
    title = {{Proceedings of the First BabyLM Workshop}},
    editor = "Charpentier, Lucas  and
      Choshen, Leshem  and
      Cotterell, Ryan  and
      Gul, Mustafa Omer  and
      Hu, Michael Y.  and
      Liu, Jing  and
      Jumelet, Jaap  and
      Linzen, Tal  and
      Mueller, Aaron  and
      Ross, Candace  and
      Shah, Raj Sanjay  and
      Warstadt, Alex  and
      Wilcox, Ethan Gotlieb  and
      Williams, Adina",
    year = "2025",
    address = "Suzhou, China",
    publisher = {{Association for Computational Linguistics}},
    url = "https://aclanthology.org/2025.babylm-main.0/",
    doi = "10.18653/v1/2025.babylm-main.0",
    ISBN = "TODO"
}

@article{WILCOX2025104650,
title = {Bigger is not always better: The importance of human-scale language modeling for psycholinguistics},
journal = {Journal of Memory and Language},
volume = {144},
pages = {104650},
year = {2025},
issn = {0749-596X},
doi = {https://doi.org/10.1016/j.jml.2025.104650},
url = {https://www.sciencedirect.com/science/article/pii/S0749596X25000439},
author = {Ethan Gotlieb Wilcox and Michael Y. Hu and Aaron Mueller and Alex Warstadt and Leshem Choshen and Chengxu Zhuang and Adina Williams and Ryan Cotterell and Tal Linzen}
}

@article{you2021child,
  title={Child-directed speech is optimized for syntax-free semantic inference},
  author={You, Guanghao and Bickel, Balthasar and Daum, Moritz M and Stoll, Sabine},
  journal={Scientific Reports},
  volume={11},
  number={1},
  pages={16527},
  year={2021},
  publisher={Nature Publishing Group UK London}
}

@article{gleitman1990structural,
  title={The structural sources of verb meanings},
  author={Gleitman, Lila},
  journal={Language acquisition},
  volume={1},
  number={1},
  pages={3--55},
  year={1990},
  publisher={Taylor \& Francis}
}

@inproceedings{feng-etal-2024-child,
    title = "Is Child-Directed Speech Effective Training Data for Language Models?",
    author = "Feng, Steven Y.  and
      Goodman, Noah D.  and
      Frank, Michael C.",
    editor = "Al-Onaizan, Yaser  and
      Bansal, Mohit  and
      Chen, Yun-Nung",
    booktitle = {{Proceedings of the 2024 Conference on Empirical Methods in Natural Language Processing}},
    year = "2024",
    address = "Miami, Florida, USA",
    publisher = "Association for Computational Linguistics",
    url = "https://aclanthology.org/2024.emnlp-main.1231/",
    doi = "10.18653/v1/2024.emnlp-main.1231",
    pages = "22055--22071"
}

@article{GILLETTE1999135,
title = {Human simulations of vocabulary learning},
journal = {Cognition},
volume = {73},
number = {2},
pages = {135-176},
year = {1999},
issn = {0010-0277},
author = {Jane Gillette and Henry Gleitman and Lila Gleitman and Anne Lederer}
}

@article{Ferguson1964BabyTI,
  title={Baby Talk in Six Languages},
  author={Charles Albert Ferguson},
  journal={American Anthropologist},
  year={1964},
  volume={66},
  pages={103-114}
}

@article{kempe2024does,
  title={{Does child-directed speech facilitate language development in all domains? A study space analysis of the existing evidence}},
  author={Kempe, Vera and Ota, Mitsuhiko and Schaeffler, Sonja},
  journal={Developmental Review},
  volume={72},
  pages={101121},
  year={2024},
  publisher={Elsevier}
}

@inproceedings{
chen2024sudden,
title={Sudden Drops in the Loss: Syntax Acquisition, Phase Transitions, and Simplicity Bias in {MLM}s},
author={Angelica Chen and Ravid Shwartz-Ziv and Kyunghyun Cho and Matthew L Leavitt and Naomi Saphra},
booktitle={The Twelfth International Conference on Learning Representations},
year={2024},
url={https://openreview.net/forum?id=MO5PiKHELW}
}

@article{gentner1978relational,
  title={{On relational meaning: The acquisition of verb meaning}},
  author={Gentner, Dedre},
  journal={{Child Development}},
  pages={988--998},
  year={1978},
  publisher={JSTOR}
}

@article{naigles1996use,
  title={The use of multiple frames in verb learning via syntactic bootstrapping},
  author={Naigles, Letitia R},
  journal={Cognition},
  volume={58},
  number={2},
  pages={221--251},
  year={1996},
  publisher={Elsevier}
}

@article{newport2004learning,
  title={{Learning at a distance I. Statistical learning of non-adjacent dependencies}},
  author={Newport, Elissa L and Aslin, Richard N},
  journal={Cognitive Psychology},
  volume={48},
  number={2},
  pages={127--162},
  year={2004},
  publisher={Elsevier}
}

@article{lidz2003infants,
  title={What infants know about syntax but couldn't have learned: Experimental evidence for syntactic structure at 18 months},
  author={Lidz, Jeffrey and Waxman, Sandra and Freedman, Jennifer},
  journal={Cognition},
  volume={89},
  number={3},
  pages={295--303},
  year={2003},
  publisher={Elsevier}
}

@article{chang-bergen-2022-word,
    title = "Word Acquisition in Neural Language Models",
    author = "Chang, Tyler A.  and
      Bergen, Benjamin K.",
    journal = "Transactions of the Association for Computational Linguistics",
    volume = "10",
    year = "2022",
    address = "Cambridge, MA",
    publisher = "MIT Press",
    url = "https://aclanthology.org/2022.tacl-1.1/",
    doi = "10.1162/tacl_a_00444",
    pages = "1--16"
}

@article{portelance2023predicting,
  title={Predicting age of acquisition for children's early vocabulary in five languages using language model surprisal},
  author={Portelance, Eva and Duan, Yuguang and Frank, Michael C and Lupyan, Gary},
  journal={Cognitive Science},
  volume={47},
  number={9},
  pages={e13334},
  year={2023},
  publisher={Wiley Online Library}
}

@inproceedings{shafiabadi-wisniewski-2025-beyond,
    title = "Beyond Surprisal: A Dual Metric Framework for Lexical Skill Acquisition in {LLM}s",
    author = "Shafiabadi, Nazanin  and
      Wisniewski, Guillaume",
    editor = "Rambow, Owen  and
      Wanner, Leo  and
      Apidianaki, Marianna  and
      Al-Khalifa, Hend  and
      Eugenio, Barbara Di  and
      Schockaert, Steven",
    booktitle = {{Proceedings of the 31st International Conference on Computational Linguistics}},
    year = "2025",
    address = "Abu Dhabi, UAE",
    publisher = "Association for Computational Linguistics",
    url = "https://aclanthology.org/2025.coling-main.443/",
    pages = "6636--6641"
}

@inproceedings{goriely-etal-2024-babble,
    title = "From Babble to Words: Pre-Training Language Models on Continuous Streams of Phonemes",
    author = "Goriely, Z{\'e}bulon  and
      Diehl Martinez, Richard  and
      Caines, Andrew  and
      Buttery, Paula  and
      Beinborn, Lisa",
    editor = "Hu, Michael Y.  and
      Mueller, Aaron  and
      Ross, Candace  and
      Williams, Adina  and
      Linzen, Tal  and
      Zhuang, Chengxu  and
      Choshen, Leshem  and
      Cotterell, Ryan  and
      Warstadt, Alex  and
      Wilcox, Ethan Gotlieb",
    year = "2024",
    booktitle = {{The 2nd BabyLM Challenge at the 28th Conference on Computational Natural Language Learning}},
    address = "Miami, FL, USA",
    publisher = "Association for Computational Linguistics",
    url = "https://aclanthology.org/2024.conll-babylm.4/",
    pages = "37--53"
}

@inproceedings{bunzeck-zarriess-2025-subword,
    title = {{Subword models struggle with word learning, but surprisal hides it}},
    author = "Bunzeck, Bastian  and
      Zarrie{\ss}, Sina",
    editor = "Che, Wanxiang  and
      Nabende, Joyce  and
      Shutova, Ekaterina  and
      Pilehvar, Mohammad Taher",
    booktitle = {{Proceedings of the 63rd Annual Meeting of the Association for Computational Linguistics (Volume 2: Short Papers)}},
    year = "2025",
    address = "Vienna, Austria",
    publisher = "Association for Computational Linguistics",
    url = "https://aclanthology.org/2025.acl-short.24/",
    doi = "10.18653/v1/2025.acl-short.24",
    pages = "286--300",
    ISBN = "979-8-89176-252-7"
}

@article{schick_function,
    doi = {10.1371/journal.pbio.3001630},
    author = {Schick, Johanna AND Fryns, Caroline AND Wegdell, Franziska AND Laporte, Marion AND Zuberbühler, Klaus AND van Schaik, Carel P. AND Townsend, Simon W. AND Stoll, Sabine},
    journal = {PLOS Biology},
    publisher = {Public Library of Science},
    title = {The function and evolution of child-directed communication},
    year = {2022},
    volume = {20},
    url = {https://doi.org/10.1371/journal.pbio.3001630},
    pages = {1-17},
    number = {5},

}

@article{golinkoff,
author = {Roberta Michnick Golinkoff and Dilara Deniz Can and Melanie Soderstrom and Kathy Hirsh-Pasek},
title ={{(Baby)Talk to Me: The Social Context of Infant-Directed Speech and Its Effects on Early Language Acquisition}},

journal = {Current Directions in Psychological Science},
volume = {24},
number = {5},
pages = {339-344},
year = {2015},
doi = {10.1177/0963721415595345},

URL = { 
    
        https://doi.org/10.1177/0963721415595345
    
    

},
eprint = { 
    
        https://doi.org/10.1177/0963721415595345
    
    

}
}

@incollection{warstadt2022artificial,
  title={What artificial neural networks can tell us about human language acquisition},
  author={Warstadt, Alex and Bowman, Samuel R},
  booktitle={{Algebraic structures in natural language}},
  pages={17--60},
  year={2022},
  publisher={CRC Press}
}

@article{liu2019roberta,
  title={{RoBERTa: A robustly optimized BERT pretraining approach}},
  author={Liu, Yinhan and Ott, Myle and Goyal, Naman and Du, Jingfei and Joshi, Mandar and Chen, Danqi and Levy, Omer and Lewis, Mike and Zettlemoyer, Luke and Stoyanov, Veselin},
  journal={arXiv preprint arXiv:1907.11692},
  year={2019}
}

@article{davies2004byu,
  title={{BYU-BNC}},
  author={Davies, Mark},
  journal={(Based on the British National Corpus from Oxford University Press). Available online at \url{http://corpus.byu.edu/bnc}},
  year={2004}
}
\end{document}